\pdfoutput=1

\documentclass[11pt]{article}

\usepackage[]{emnlp2021}
\usepackage{amsthm}
\usepackage{amsmath,amsfonts,bm}
\makeatletter
\newtheorem*{rep@theorem}{\rep@title}
\newcommand{\newreptheorem}[2]{%
\newenvironment{rep#1}[1]{%
 \def\rep@title{#2 \ref{##1}}%
 \begin{rep@theorem}}%
 {\end{rep@theorem}}}
\makeatother

\newtheorem{definition}{Definition}[section]

\newreptheorem{theorem}{Theorem}

\newcommand{\RNum}[1]{\uppercase\expandafter{\romannumeral #1\relax}}

\newcommand{\G}{\mathcal{G}}

\newcommand{\V}{\mathcal{V}}
\newcommand{\E}{\mathcal{E}}

\newcommand{\R}{\mathcal{R}}

\usepackage{mathtools}

\newcommand{\xhdr}[1]{{\noindent\bfseries #1}.}
\newcommand{\cut}[1]{}

\newcommand{\removelatexerror}{\let\@latex@error\@gobble}

\def\eqref#1{Eq.~\ref{#1}}

\def\1{\bm{1}}

\DeclareMathAlphabet{\mathsfit}{\encodingdefault}{\sfdefault}{m}{sl}
\SetMathAlphabet{\mathsfit}{bold}{\encodingdefault}{\sfdefault}{bx}{n}

\usepackage{xcolor,color}
\usepackage{times}
\usepackage{latexsym}
\usepackage{tikz}
\usepackage{bbm}
\usepackage{booktabs} 
\usepackage{enumitem}
\usepackage{url}
\usepackage{multirow}
\usepackage{amsmath}
\usepackage{amsfonts}
\usepackage{graphicx}
\usepackage{footnote}

\usepackage[T1]{fontenc}

\usepackage[utf8]{inputenc}

\usepackage{microtype}

\usepackage{xifthen}
\usepackage{graphicx}
\usepackage{subfigure}
\usepackage{lipsum}
\usepackage{mathtools}
\usepackage{cuted}
\usepackage{times}
\usepackage{url}
\usepackage[makeroom]{cancel}
\usepackage{latexsym}
\usepackage{pifont}
\usepackage{amsmath}
\usepackage{footnote}
\usepackage{graphicx}
\usepackage{amsfonts}
\usepackage{bm}
\usepackage[mode=text]{siunitx}
\usepackage{thmtools, thm-restate}
\usepackage{amsmath, amsthm, amssymb}
\usepackage{etoolbox}
\usepackage{multirow}
\usepackage{caption}
\usepackage{amsmath,amsfonts,amssymb}
\usepackage{mathrsfs}
\usepackage{tabu}
\usepackage{adjustbox}
\usepackage{color}
\usepackage{xcolor}
\usepackage{colortbl}
\usepackage[normalem]{ulem}
\usepackage{amsthm}

\theoremstyle{remark}
\newtheorem{remark}{\textbf{Observation}}

\theoremstyle{definition}
\definecolor{emerald}{rgb}{0.31, 0.78, 0.47}
\definecolor{lightcoral}{rgb}{0.94, 0.5, 0.5}
\definecolor{gold(web)(golden)}{rgb}{1.0, 0.84, 0.0}
\definecolor{lightcornflowerblue}{rgb}{0.6, 0.81, 0.93}
\definecolor{Gray}{gray}{0.9}

\newcommand{\algoname}{\textsc{Neural Path Hunter}}
\newcommand{\algonameshort}{\textsc{NPH}}

\usepackage{hyperref}
\title{Neural Path Hunter: Reducing Hallucination in Dialogue Systems via Path Grounding}

\author{
  Nouha Dziri\thanks{\quad Corresponding author.}$^*$, Andrea Madotto$^\dagger$, Osmar Zaiane$^{*\S}$, Avishek Joey Bose$^\ddagger$\\
  $^*$University of Alberta, $^\ddagger$Mila, McGill University, $^\S$Canada CIFAR AI Chair  \\
  $^\dagger$The Hong Kong University of Science and Technology \\
  \texttt{dziri}\texttt{@cs.ualberta.ca}
  }

\begin{document}
\maketitle
\begin{abstract}
Dialogue systems powered by large pre-trained language models exhibit an innate ability to deliver fluent and 
natural-sounding responses.
Despite their impressive 
performance, these models are fitful and can often generate factually incorrect statements impeding their widespread adoption. In this paper, we focus on the task of improving faithfulness and reducing hallucination of neural dialogue systems to known facts supplied by a Knowledge Graph (KG). We propose \algoname \ which follows a generate-then-refine strategy whereby a generated response is amended using the KG. \algoname \ leverages a separate token-level fact critic to identify plausible sources of hallucination followed by a refinement stage that retrieves correct entities by crafting a query signal that is propagated over a $k$-hop subgraph.\cut{Our proposed model can easily be applied to any dialogue generated responses without retraining the model.} We empirically validate our proposed approach on the OpenDialKG dataset \citep{moon2019opendialkg} against a suite of metrics and report a relative improvement of faithfulness over dialogue responses by $20.35\%$ based on FeQA \cite{durmus2020feqa}.  The code is available at \url{https://github.com/nouhadziri/Neural-Path-Hunter}.
\end{abstract}

 \section{Introduction}
\cut{Conversation within a dialogue can be thought of
as a kind of action being performed by the speaker within each turn or utterance \cite{jurafsky2018speech}. For instance, an utterance can be a directive whereby the speaker requests the addressee to perform actions e.g. asking, ordering, or forbidding. However, each utterance is not independent of one another but is instead grounded within a larger dialogue context known to both parties \cite{sordoni2015neural, serban2016building, dziri2019augmenting}}
Conversation within a dialogue can be thought of as an exchange of utterances between two speakers. Each utterance is not independent of one another but is instead grounded within a larger dialogue context known to both parties \cite{jurafsky2018speech, sordoni2015neural, serban2016building, dziri2019augmenting}. Indeed, if a response to an utterance fails to be faithful to some given knowledge---i.e. by producing false information---it is uninformative and runs the risk of jeopardizing the entire enterprise of conversation. More precisely, this means that in addition to being fluent, topical, and grammatical,  utterances within a dialogue must also be \textit{factually correct}.

\begin{figure}[t]
    \centering
    \includegraphics[width=0.99\linewidth]{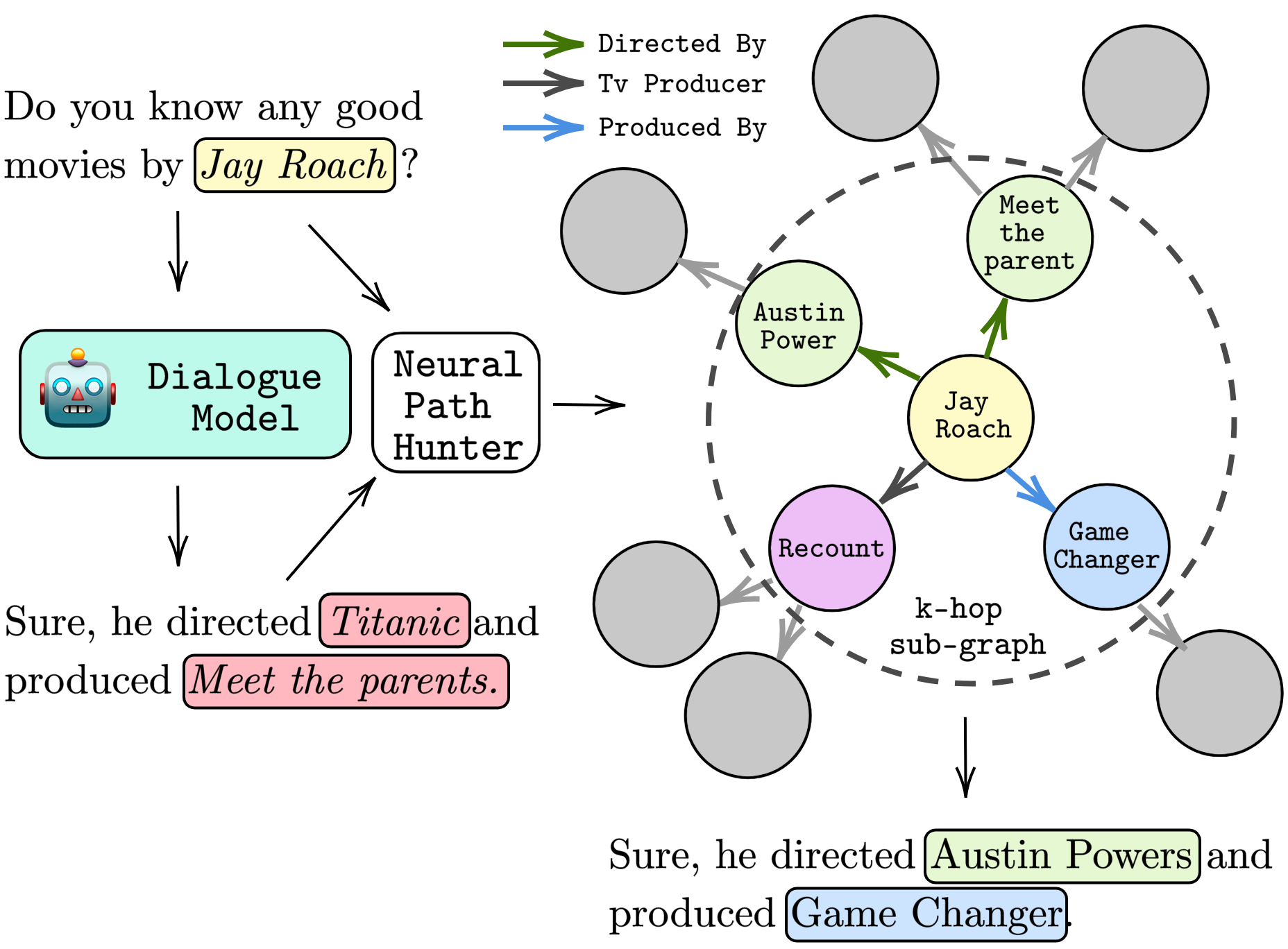}
    \caption{$\algoname$ overview.}
    \label{fig:intro_approach_overview}
    \vspace{-15pt}
\end{figure}

The faithfulness of responses is of principal importance when designing dialogue systems that are grounded using auxiliary knowledge such as Knowledge Graphs (KG). Despite maintaining plausible general linguistic capabilities, dialogue models are still unable to fully discern facts and may instead hallucinate factually invalid information. 
Moreover, empirical evidence for hallucination in Language Models (LM) runs contrary to known studies that these large models are capable of recalling factual knowledge, e.g. entities and relations in a KG, \cite{roberts2020much,petroni2019language}. This suggests that this inherent lack of controllability may be remedied by leveraging external oracle knowledge. However, existing approaches to knowledge grounding often suffer from a source-reference divergence problem whereby the reference contains additional factual information and simply training on the reference is insufficient to guarantee faithfulness \cite{wiseman2017challenges,parikh2020totto,tian2019sticking}. Consequently, ensuring the faithfulness of knowledge grounded dialogue systems---via precise alignment of the source and reference---remains an open challenge.

\xhdr{Present Work}
In this work, we focus on addressing the open problem of hallucination of factually invalid statements in knowledge grounded dialogue systems where the source of knowledge is a KG. We first identify prominent modes of hallucination by conducting a systematic human study on generated responses which reveals one major source of hallucination as the (mis)-use of wrong entities to describe factual content \cite{kryscinski2020evaluating}, a problem that persists when naively applying language models in dialogue systems. \cut{Concretely, we seek to design KG-grounded dialogue systems capable of generating utterances that are both semantically relevant given the conversation history \textit{and} are supported within a provided KG.} 

To enforce faithfulness to the misattribution of entities in grounded dialogue systems, we introduce \algoname \ ($\algonameshort$), a module that operates on hallucinated responses. \algonameshort \ follows a generate-then-refine approach by augmenting conventional dialogue generation with an additional refinement stage enabling the dialogue system to correct potential hallucinations by querying the KG. \algonameshort \ grounds dialogue generation by constraining the flow of conservation to be supported by a valid path on the KG. To do so, the module combines a token-level hallucination critic that masks out entities of concern in an utterance, followed by a pre-trained non-autoregressive LM which prescribes contextual representations for each masked entity. This is then fed sequentially to an autoregressive LM to obtain output representations. These output representations can then be used to efficiently launch a query on the KG---effectively modelling dialogue as a signal being propagated on a local $k$-hop subgraph whereby locality is enforced through the conversation history---returning factually correct entities. Our proposed approach is applicable to any generated response whenever an available KG is provided and works without further fine-tuning. The high-level overview of our proposed approach is outlined in Fig. \ref{fig:intro_approach_overview} and exemplar machine-generated responses post-refinement are presented in Table \ref{tab:examples_fixed} in~\S\ref{app:samples}.
 Our main contributions are summarized as follows:
\begin{itemize}[noitemsep,topsep=0pt,parsep=0pt,partopsep=0pt,label={\large\textbullet},leftmargin=*]

\item We conduct a comprehensive human study on hallucinations generated by state-of-the-art dialogue systems which reveals that the main mode of hallucinations is through the injection of erroneous entities in generated responses.

\item We propose \algoname, which leverages facts supplied by a KG to reduce hallucination in any machine-generated response. 

\item We empirically demonstrate that \algoname \ substantially reduces hallucinations in KG-grounded dialogue systems with a relative improvement of $20.35\%$ in FeQA, a QA-based faithfulness metric \cite{durmus2020feqa}, and an improvement of $39.98 \%$ in human evaluation. 

\end{itemize}

\label{app:samples}
 \enlargethispage{2em}
 \begin{table*}[htbp]
 \centering
    \begin{tabu}to\linewidth{@{}X[1.2,l]X[0.2,l]X[6,l]@{}}
    \toprule
    \textbf{\small History} & $A$: & Do you know the book The Witches?\\
      & $B$: & The Witches is written by Roald Dahl. He also wrote The Champion of the World.\\
      \midrule
      \textbf{\small GPT2-KG} & $A_{gen}$ & Yes he did. He also wrote \colorbox{lightcoral}{The Time Machine} and \colorbox{lightcoral}{The Invisible Man}. \\
    \midrule
     \textbf{\small Gold knowledge}
    &  $T_1$: & [Charlie and the Chocolate Factory, written by, Roald Dahl] \\
    &  $T_2$: & [Charlie and the Chocolate Factory, has genre, Fantasy]\\
        \midrule
     \textbf{\small Top-5 Paths}
    & $T'_1$: & [The BFG, written by, Roald Dahl]
\\
    & $T'_2$: & [Charlie and the Chocolate Factory, written by, Roald Dahl]
 \\
    & $T'_3$: & [You Only Live Twice, written by, Roald Dahl] \\
    & $T'_4$: & [James and the Giant Peach, written by, Roald Dahl]
 \\
    & $T'_5$: & [Tales of the Unexpected, TV regular appearance, Roald Dahl]
    \\   \midrule
       \textbf{\small  $\algonameshort$ response} & $A_{fix}$ &Yes he did. He also wrote \colorbox{emerald}{The BFG} and \colorbox{emerald}{Charlie and the Chocolate Factory}. \\
    \bottomrule

    \end{tabu}
    \caption{A selected response based on a GPT2-KG test response before and after applying \algoname. The span of texts highlighted in red indicate the hallucinated entity mentions whereas the ones highlighted in green indicate the retrieved correct entity mentions. }
    \label{tab:examples_fixed}
    \end{table*}
\section{Hallucination in KG-grounded Dialogue Systems}
We consider the task of generating factual and grounded dialogue when presented with auxiliary structured knowledge. In particular, we focus on factoids taken from multi-relational graphs $\G = (\V, \E, \R)$, termed Knowledge Graphs (KG). Each KG consists of a set of directed edge triples $t = \langle \texttt{[SBJ]}, \texttt{[PRE]}, \texttt{[OBJ]} \rangle$, where $\texttt{[SBJ]}, \texttt{[OBJ]} \in \V$ are nodes denoting subject and object entities and $\texttt{[PRE]}\in \R$ is a predicate that can be understood as a relation type. Broadly speaking, we say that a neural dialogue system is guilty of hallucinating whenever it generates a factual sentence that is not supported by a valid path in a $k$-hop subgraph $\mathcal{G}^k_c \subset \mathcal{G}$ of the original KG anchored around a context entity $c$. 


As a starting point for our investigation, we study the various types of hallucinations a model may inject into an otherwise satisfactory response. Specifically, we explore the circumstances under which LMs are likely to exhibit unfaithful behaviour through misappropriation of entities (e.g. Barrack Obama was the President of Canada). Inspired by \citep{maynez2020faithfulness} for KG-grounded dialogue systems we hypothesize---among other possible mechanisms---hallucination can take form as either \textit{intrinsic} or \textit{extrinsic} to the provided KG. 

\begin{definition}[Extrinsic Hallucination]
\label{def:extrinsic_hallucination}
An extrinsic hallucination corresponds to an utterance that brings a new span of text that does not correspond to a valid triple in $\mathcal{G}^k_c$. 

\end{definition}
From the perspective of definition~\ref{def:extrinsic_hallucination}, an utterance that might be partially faithful is still guilty of hallucination if there exists any injection of knowledge not authentically captured in $\mathcal{G}^k_c$. Despite this, external hallucinations can often be easier to identify due to their egregious nature. For example, the dialogue sample in Fig. \ref{fig:intro_approach_overview} contains an external hallucination as the entity in question ``Jay Roach'' did not direct the movie ``Titanic'' and it is not supported within the $1$-hop subgraph. On the other hand, the generated response may identify the correct set of entities but make false claims about their relationship which leads to the following definition. 

\begin{definition}[Intrinsic Hallucination]
\label{def:intrinsic_hallucination}
An intrinsic hallucination corresponds to an utterance that misuses either $\texttt{[SBJ]}$ or $\texttt{[OBJ]}$ in $\mathcal{G}^k_c$ such that there is no direct path between the two entities. 
\end{definition}
Intrinsic hallucinations inject false information by condensing information from the KG in a wrong way. For instance, claiming that ``Jay Roach'' produced ``Meet the Parents'' is an incorrect association of the true relationship between these entities.

To ascertain the degree to which KG-grounded dialogue systems hallucinate and the nature of these hallucinations, we conduct a systematic evaluation by soliciting human judgement. We first fine-tune a LM on the OpenDialKG dataset \cite{moon2019opendialkg} which contains a turn-based dialogue between two speakers on extracted triples from a known KG. The sequential nature of such turn-based dialogues grounded via extracted KG-triples effectively renders the entire conversation as a path traversed on the KG (see \S\ref{appendix:opendial_kg} for dataset details). 

\subsection{Modes of Hallucination}
\label{sec:hallu}

\xhdr{Experimental Protocol}
As a demonstrative example, we use a pre-trained GPT-2 model \cite{radford2019language} as the backbone of a neural dialogue system. To fine-tune GPT2, we concatenate the dialogue history, the KG-triples $\langle \texttt{[SBJ]}, \texttt{[PRE]}, \texttt{[OBJ]} \rangle$ and the ground truth response and then train the model to predict the next word in the response. To explore the effect of different decoding strategies and their impact in injecting hallucinations, we sample $300$ responses from each decoding approach. We investigate greedy search, beam search, nucleus sampling \cite{holtzman2019curious} and top-$k$ sampling \cite{radford2019language} as representative decoding strategies. 

\begin{table}[t]
\begin{small}
\centering
\begin{tabular}{ l | c c c | c | c }
 \hline
  \multirow{2}{*}{\small \textbf{GPT2-KG}} & \multicolumn{3}{c|}{{\small \textbf{Hallucination}}} & \multirow{2}{*}{{\small \textbf{Faith.}}} & \multirow{2}{*}{{\small \textbf{Gen.}}}  \\
  & \small \textbf{Ex} & \small \textbf{In} & \small \textbf{B} & \\
 \hline
 {\small Greedy} & {\small \textbf{17.66} } &  {\small \textbf{2.00}}&{\small \textbf{1.66}} & {\small \textbf{ 69.00}} & {\small 9.66 }\\
 {\small Beam Search} & {\small 18.33} & {\small 3.33  } & {\small 4.00} & {\small 68.00} & {\small 6.33} \\
 {\small Nucleus 0.9} & {\small 25.33} & {\small 4.00 } &{\small 2.33} & {\small 64.66}& {\small \textbf{3.66}}  \\
  {\small Nucleus 0.5} & {\small 23.33} & {\small 5.33} & {\small 4.33} & {\small 59.90} & {\small 7.00 }  \\
  {\small Top20} & {\small 28.33} & {\small 7.00}& {\small 5.00} & {\small 55.00} & {\small 4.66} \\ \hline
\end{tabular}
\caption{{Human assessment of random $1500$ GPT2 dialogue responses generated using OpenDialkg. ``Ex", ``In" and "B" mean extrinsic, intrinsic, and both hallucinations respectively. Each cell shows the mean percentage of responses with a specific dialogue property (see~\S\ref{detailed_mode_hall} for confidence intervals). 
}}
\label{tab:human_assess_gpt2}
\end{small}
\vspace{-5mm}
\end{table}

For each dialogue sample, we crowd-source human judgement by soliciting evaluations from $3$ different annotators from Appen\footnote{\url{https://appen.com/}}, a high-quality annotation platform. Each annotator is tasked to first identify the presence of hallucination in the generated response when provided the dialogue history and KG triples. 
For samples where hallucination is present, we further ask the human annotators to identify whether the hallucination is extrinsic, intrinsic or both. If the response is not hallucinated, we ask them whether the response is faithful (i.e., supported by the triples) or generic (e.g., ``I don't know about that'').\cut{Finally, the human annotators are also tasked to assess the general fluency and coherence of the generated response agnostic of hallucinations.} The results of the human assessment are shown in Table \ref{tab:human_assess_gpt2}. Overall, we report the average Krippendorf's alpha coefficient to be $0.72$ on the annotator responses to the different questions which indicates high agreement. Using Table~\ref{tab:human_assess_gpt2}, we make the following key observations:

\begin{remark}
\label{obs:one}
Humans notice most hallucinations in KG-grounded dialogue systems are extrinsic.
\end{remark}

\begin{remark}
\label{obs:two}
A hallucination occurs the least in dialogue responses generated using a greedy decoding scheme. Conversely, top-$k$ sampling results in the highest hallucination percentage ($40.33\%$). 
\end{remark}

\begin{remark}
\label{obs:three}
Increased diversity in response generation ---i.e.(less generic), is positively correlated with an increase in hallucination e.g. Nucleus=$0.9$. 

\end{remark}
\cut{
\begin{remark}
\label{obs:four}
Responses from all models tend to be highly relevant and fluent, which reflects the ability of pre-trained LMs in generating human-like responses.
\end{remark}
}

Observation~\ref{obs:one} indicates that the dominant mode of hallucination for all decoding strategies in KG-grounded dialogue systems is extrinsic rather than intrinsic. In fact, we find that in the OpenDialKG dataset, $54.80\%$ of the responses contain extra entity mentions that are not supported by either $\mathcal{D}$ or $\mathcal{G}^1_c$ which may partially explain empirical observations. Observation~\ref{obs:two} suggests that the model---when conditioned on factual knowledge---often assigns the highest probability mass to the correct response and sampling based on other distributions (e.g. top-$k$) invites hallucination in the generation process---a fact also observed in language modelling \cite{keskar2019ctrl}.  Observation~\ref{obs:three} suggests an implicit trade-off between the different goals of response generation whereby improving the diversity of response can negatively impact its faithfulness. This reveals that in certain cases responses might be originally faithful to $\mathcal{G}^k_c$ but increasing diversity encourages the model to hallucinate. \cut{Lastly, observation~\ref{obs:four} shows that responses are often coherent with respect to the history and are fluent even though they maybe hallucinated in some cases. Also, this reveals that in certain cases responses might be faithful to $\mathcal{G}^k_c$ but not necessarily relevant to the dialogue history.} In light of these important observations, the main goal of this paper is not necessarily to advance state-of-the-art decoding methods but instead to instrument an efficient technique to identify hallucinations as well as retrieve the correct entities from the KG. 


\cut{
\xhdr{Dataset} We used OpenDialKG \cite{moon2019opendialkg}, a crowed-sourced English dialogue dataset where two workers are paired together to chat about a certain topic. The first speaker is asked to initiate the conversation about a given entity and the second speaker is given a set of facts extracted from an existing knowledge graph (KG), Freebase \cite{bast2014easy}, from which they have to select relevant facts to form a factual response. Those facts represent paths in the KG that are either 1-hop or 2-hop from the initial entity. Once the second speaker sends a response, the first speaker continue discussing the topic engagingly and new multi-hop facts from the KG are presented to the second speaker, including paths initiating from entities discussed in the previous turn. The conversation can be regarded as traversing multi paths in the KG. However, not all utterances within the same dialogue are grounded on facts from the KG. The second speaker can choose not to select paths from the KG to form an answer.
Overall, the dataset consists of four domains: movie, music, sport and book where each second speaker's utterance is annotated with paths from the KG. The KG corresponds to a large subgraph extracted from Freebase with $\sim1.2$M triples (subject, predicate, object), $\sim101$k distinct entities and 1357 distinct relations. No official
split is provided and thus we randomly split the
dataset in 80/10/10 for the train/valid/test, respectively.
The data consists of 61778 train, 7933 valid and 7719 test.  (Provide numbers about the ratio of examples in the train, test, dev contains entities that do not appear in the gold kg triples.)

For fine-tuning GPT2, we concatenate the dialogue history, the fact triples (subject, predicate, object) and the response and train the model to predict the next word in the sequence. \texttt{BOS} \texttt{[SPK1]} utt1 \texttt{[SBJ]} subject \texttt{[PRE]} predicate \texttt{[OBJ]} object \texttt{[speaker2]} utt2 \texttt{EOS}.

To explore the effect of different decoding strategies on generating hallucination, we randomly sample 300 responses from each decoding approach. We explore greedy search, beam search, nucleus sampling \cite{holtzman2019curious} and top-k sampling \cite{radford2019language}. We crowd-sourced human assessment from Appen \footnote{\url{https://appen.com/}}, a high-quality annotation platform, where workers were trained for the task before starting the evaluation process. If the workers fail to achieve at least 70\% accuracy in answering the different evaluation questions, they won't be allowed to start evaluating. We have trained workers who are fluent in English. For each dialogue example, we collect judgments from 3 different workers who were presented with a dialogue history of length 3,  knowledge triples that are dedicated to produce the next response and the generated response. We ask the annotators to first identify whether the response is hallucinated w.r.t the knowledge triples, if yes, a followup questions that's inspired from \citet{maynez2020faithfulness} would be to identify the hallucinated spans in the text and to specify whether it's an intrinsic hallucination or an intrinsic hallucination. An intrinsic hallucination corresponds to a response that misuses information like named entities from the knowledge triples and the history resulting in an unfaithful sentence such as . Examples can be seen in Table \ref{}. An extrinsic hallucination corresponds to a response that brings new facts that are not supported by the knowledge triples. The annotators are further asked whether the response exploits the knowledge triples in forming the response and this to ensure this point: if the response was not hallucinated, we want to investigate whether the response was factual by using the knowledge (in this case the response is considered faithful) or if it was just generic like "I don't know much about Jay Roach's movies."

The last two questions are about relevance and fluency. We ask humans whether the response was coherent with the dialogue history even if it's faithful and whether it's fluent and does not suffer from grammatical issues.  
}

\cut{
\begin{itemize}[noitemsep,topsep=0pt,parsep=0pt,partopsep=0pt,label={\large\textbullet},leftmargin=*]
    \item Hallucination happens the least in dialogue responses that are generated greedily. This can be explained by the ability of a very well-trained model to assign a high probability weight to the correct factual response, i.e., if we sample the highest probable word at every time step, it will result in a more factual/correct response. When conditioned on some factual knowledge, the model is more confident about the answer. On the other hand, by focusing on sampling from the head of the distribution instead of the most probable word at each time step, we mistrust the model and thus hallucination happens. 
    This is a similar to the observation made in \cite{keskar2019ctrl} for the task of language modelling. 
    
    \item we can notice the more diverse the response becomes (less generic), the more hallucinated it becomes. This can be observed with Nucleus=0.9 which was designed to sample from the smallest set of words with a cumulative probability  above a threshold.
    
    \item Sampling randomly from the TopK words of the distribution results in the highest hallucination percentage ($37.33\%$)
    
    \item Humans notice that most of hallucinations are extrinsic. extrinsic hallucination is due to the knowledge-response divergence, to the large pre-trained model that has learnt many knowledge during training and also to the decoding approaches (sampling from the head of the distribution)
    \item For each generated response, humans were asked to identify the span of text that is hallucinated. By going through all the identified spans, we noticed that the majority of hallucinated spans are related to named entities such as name of a person, movie, etc. And also hallucination can be related to movie genres, professions (Not sure what do we call this).
    \item this suggests that there is a trade-off between the factuality and different goals of response generation. If the intended goal of response generation is diversity, the model will sacrifice faithfulness.
\end{itemize}
}

\section{Neural Path Hunter}
\label{sec:nph}
 \begin{figure}[t]
     \centering
     \includegraphics[width=0.9\linewidth]{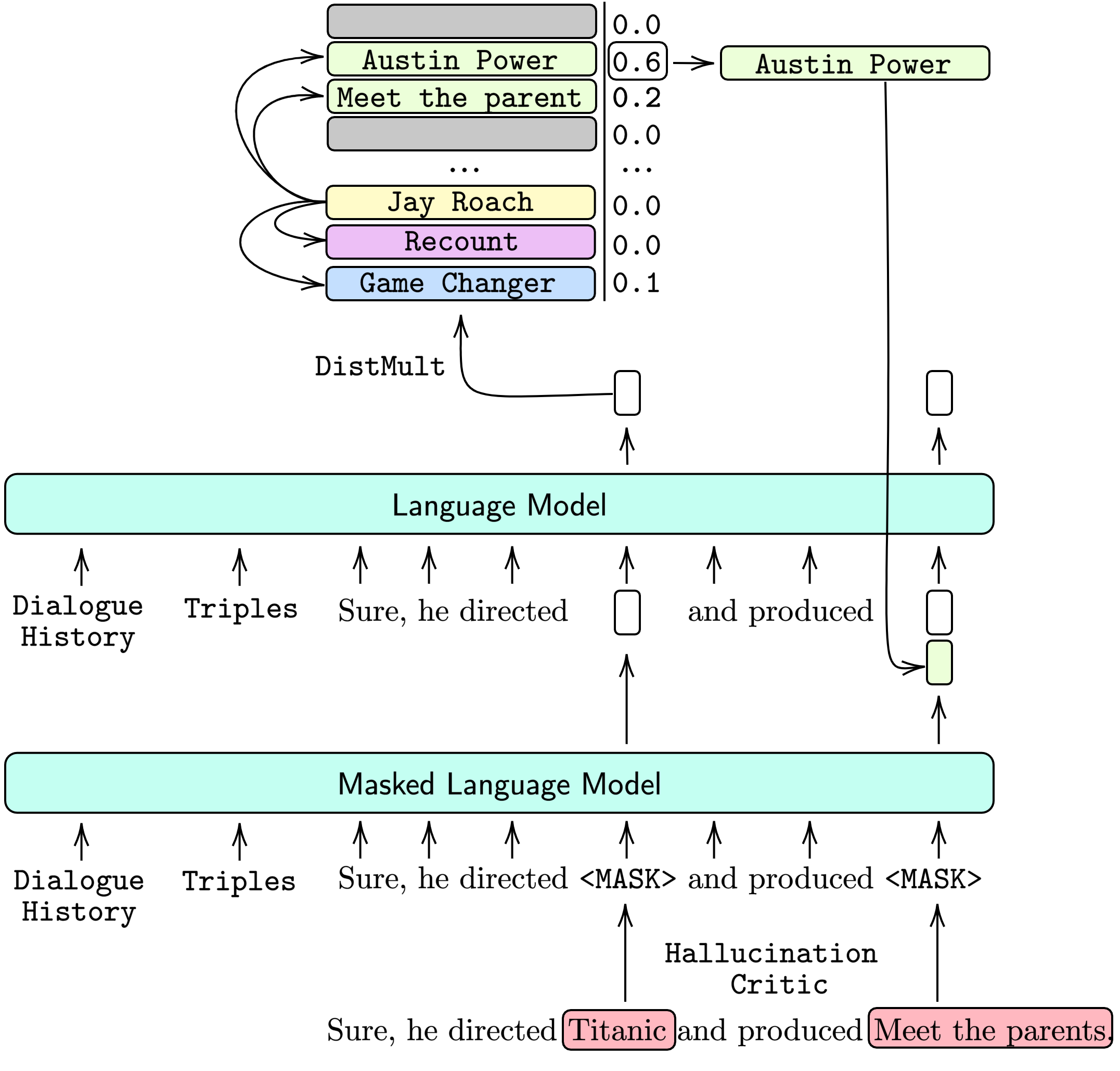}
     \caption{Entity Mention Retriever architecture.}
     \vspace{-15pt}
     \label{fig:diagram_emr}
 \end{figure}
 
We seek to design a dialogue refinement system capable of fixing generated utterances such that they are semantically relevant given the conversation history \textit{and} supported within a provided KG. To do so, we introduce \algoname \ (\algonameshort)\ a refinement strategy that can be easily applied to any generated response without retraining the model. \algonameshort \ is composed of two modules: A token-level hallucination critic and an entity mention retriever. The first module flags and masks out hallucinated entities in an existing response and can be trained offline. The second module accepts masked representations identified by the critic and builds contextual representation of these problematic tokens which are then used to retrieve more faithful entities by running a query over $\mathcal{G}^k_c$. We assume the local $k$-hop subgraph is either provided or extracted based on the dialogue history. The following sections describe the data preparation, training, and inference procedures for these submodules.

\subsection{Problem Formulation}
Each instance in the dataset is composed of a dialogue history $\mathcal{D} = (x_1, \dots, x_n)$, a set of $j$ triples at turn $n$, $\mathcal{K}_{n}= (t_1, t_2, \dots t_j)$ which together with $\mathcal{D}$ must be used towards generating the response $\bar{x}_{n+1}$. Here, each individual triple $t_i = \langle \texttt{[SBJ]}, \texttt{[PRE]}, \texttt{[OBJ]} \rangle$ is extracted from a provided KG. Thus, the task is to generate a response $\bar{x}_{n+1}$ that is faithful to a non-empty subset $M_n \subset \mathcal{K}_n$ ---i.e., it can optionally talk about a few triples but not none. Specifically, the response $\bar{x}_{n+1}$ may contain entity mentions $m_i \in \mathcal{V}$ which indicates a factual response that potentially needs to be refined using \algonameshort. 
For our purposes, it is most convenient to represent each mention as a tuple of three elements that indicates the beginning of the mention at position $m^b_i$ and the end at position $m^e_i$. In other words, we represent an entity mention $m_i$ as $m_i = (m_i, m^b_i, m^e_i)$. These entity mentions may not be faithful at all if they do not belong to either a $\texttt{[SBJ]}$ or $\texttt{[OBJ]}$ in $M_n$ (extrinsic hallucination) or they could inject false relationships between mentions via an unsupported path in $\mathcal{G}^k_c$ by incorrectly utilizing a $\texttt{[PRE]}$ (intrinsic hallucination). We target and correct these unfaithful entities through retrieval over $\mathcal{G}^k_c$ in ~\S\ref{entity_mention_retriever}.

\subsection{Token-level hallucination critic}
\label{critic}
To enforce faithfulness via refinement, we first identify the exact sources of hallucination in a given response. Based on the findings of human judgement in Tab.\ref{tab:human_assess_gpt2} and ~\S\ref{sec:hallu}, we find hallucination errors in a dataset like OpenDialKG are often associated with entity mentions such as names of people, movies titles, locations, etc. To flag entities of concern, we design a token-level hallucination critic $C$ that consumes $\mathcal{D},\mathcal{K}_n, \bar{x}_{n+1}$ and outputs the set of hallucinated entity mentions $M_c$. To train $C$, we choose to cast the problem as a sequence labelling task where a binary label is predicted at each word position. As there is no labelled training data available for this task, we create a synthetic dataset consisting of ground truth dialogue samples and corrupted negative samples. We explore two corruption processes that convert a regular clean ground-truth response $x_{n+1}$ to its corresponding hallucinated one $\hat{x}_{n+1}$ based on the type of hallucination we might expect to encounter ---i.e. extrinsic and intrinsic.

\begin{enumerate}[noitemsep]
  \item \xhdr{Extrinsic Negatives} We replace each $m_i$ in $x_{n+1}$ with entities of the same type (e.g., person, location, etc...) but crucially not within $\mathcal{G}^k_c$ and the dialogue history $\mathcal{D}$. 
  \item \xhdr{Intrinsic Negatives} We simply swap every pair $\texttt{[SBJ]}$ and $\texttt{[OBJ]}$ in $x_{n+1}$. For example, the response ``Crescendo was written by Becca Fitzpatrick'' $\to$ ``Becca Fitzpatrick was written by Crescendo'' results in an intrinsic hallucination as in this case $\texttt{[PRE]}$ is not bi-directional.
\end{enumerate}

Overall, we apply a $60\%/40\%$ split of extrinsic versus intrinsic corruption strategies to the original train OpenDialKG to obtain a synthetic dataset to train $C$ which is taken to be a pre-trained LM that is then fine-tuned on this binary classification task.


\subsection{Entity Mention Retriever}
\label{entity_mention_retriever}
An overview of the Entity Mention Retriever is depicted in Fig. \ref{fig:diagram_emr}.
Having identified entities of concern in $\bar{x}_{n+1}$, we now wish to craft a query that can be efficiently run over $\mathcal{G}^k_c$. To do so, we model the generated response $\bar{x}_{n+1}$ as a signal being propagated over $\mathcal{G}^k_c$ which serves to capture the highest probability paths starting from the context node $c$ the conversation may take if it was faithful. The context node $c$ is extracted from ground truth triples available in the dataset and or $\mathcal{D}$. In order to run an effective query over $\mathcal{G}^k_c$, it is critical that the representation of all flagged $m_i \in M_c$ and edge triples $\mathcal{E} \in \mathcal{G}^k_c$ are in the same representation space. Inspired by the Cloze task \cite{taylor1953cloze}, we obtain contextual representations of all $m_i$'s identified by the critic by first masking them out before using a Masked Language Model (MLM).
Operationally, we feed $\mathcal{D}$, $\mathcal{K}_n$, as well as the flagged set of entities to obtain contextual hidden state representations:
\begin{equation}
 H = \text{MLM}({\mathcal{D},\mathcal{K}_n,M_c})
\end{equation}
As the MLM may return multiple hidden $d$-dimensional state representation for each $m_i \in M_c$, we simply apply a pooling operation to obtain a single representation for each entity ---i.e. $h_i = \text{MaxPool}(h_b, h_e)$. To obtain the actual query $q_i$, we use an autoregressive LM  which iteratively consumes an order dependent representation of $h_i$ given by applying a learnable projection map $W: \mathbb{R}^{2d} \to \mathbb{R}^d$ to a concatenation of the current hidden state and the retrieved entity embedding $e_{i-1}$ using previous query $q_{i-1}$ as shown in Fig. \ref{fig:diagram_emr}, 
\begin{equation*}
    q_i = \text{LM}(W(\text{concat}[e_{i-1}, h_i])),
\end{equation*}

\xhdr{KG-Entity Memory}
Viewed another way, each $q_i$ can be interpreted as a relation embedding for the masked position in $\bar{x}_{n+1}$. To effectively query $\mathcal{G}^k_c$, we must also represent all nodes in the same embedding space as $q_i$ and in doing so effectively build a representation of $\mathcal{G}^k_c$ which we call KG-Entity Memory. We explore two approaches towards this goal. The first uses the final hidden layer of a pre-trained GPT2 to obtain initial embeddings for each node in $\mathcal{G}^k_c$ \footnote{Actually, GPT2 returns word piece representations and we use a pooling operation to get a single representation.}. Our second approach uses CompGCN \cite{vashishth2019composition}, which is a Graph Convolutional Network \cite{kipf2016semi} purposely built for multi-relational data. We initialize the CompGCN network offline with GPT2 embeddings for all entities and relations in the full graph $\mathcal{G}$ before running a few rounds of message passing by optimizing for a standard relation prediction objective. Both approaches to KG-Entity memory embeddings can be further updated during training. Finally, to retrieve the correct entity for query $q_i$, we simply use a scoring function $s$ to score every KG-Entity memory triple in $\mathcal{G}^k_c$  ---i.e. $t_i = \langle c, q_i, \texttt{[OBJ]} \rangle$. The retrieved entity is the $\texttt{[SUB]}$ or $\texttt{[OBJ]}$ that achieves the highest score.

\subsection{Training the Entity Mention Retriever}
\label{sec:training}
To train the Entity Mention Retriever, we augment the conventional maximum likelihood objective with an additional contrastive loss $\mathcal{L_{\text{NCE}}}$ that encourages faithful retrieval. In particular, we use Noise Contrastive Estimation (NCE) \cite{gutmann2010noise} which forces the Entity Mention Retriever to learn a scoring rule such that $s(t_i) > s(t_i'), \forall t_i \in \E, t_i' \in \bar{\E}$ where $t_i=\langle c, q_i, \texttt{[OBJ]} \rangle$ is the edge-triple based on KG-entity memory and $t_i'=\langle c, q_i, \texttt{[OBJ]}^{-} \rangle$ is a negative sample where $\texttt{[OBJ]}^{-}$ \footnote{or $\texttt{[SUB]}^{-}$ if $c$ is an object} is sampled from a corruption distribution over edge triples $\bar{\E}$ \textit{not} in $\mathcal{G}^k_c$. To compute $\mathcal{L_{\text{NCE}}}$, we draw $n$ negative samples uniformly over all entities for each query $q_i$.
\begin{align}
\mathcal{L_{\text{NCE}}} &= -\log \left (s(t) \right )  - \log \bigg (s(t) +  \sum_{j=1}^{n} s(t') \bigg ). \nonumber
\end{align}
At training time, we use teacher forcing \cite{williams1989experimental}; first, we mask out all entity mentions within the gold response $x_{n+1}$, get their representations through a MLM and provide the ground truth entity mention concatenated with $h_i$ at each time step in the LM. For the scoring function, we use DistMult \cite{wang2014knowledge} due to its simplicity in the absence of known structure over the modified triples e.g. translation, rotation, which are exploited in other popular scoring functions for KGs. By optimizing $\mathcal{L_{\text{NCE}}}$, we encourage the model to leverage the dialogue history, the position of the masked entity in $x_{n+1}$, and the $k$-hop subgraph to identify more faithful entities that are relevant to the conversation history. To train the Entity Mention Retriever, we thus jointly optimize $\mathcal{L}_{\text{NCE}}$ and $\mathcal{L}_{\text{MLE}}$ for the main language modelling task, 
\begin{equation}
\mathcal{L} = \mathcal{L_{\text{MLE}} + \lambda \mathcal{L_{\text{NCE}}}}.
\label{eq:loss}
\end{equation}

\section{Experiments}
We evaluate the ability of \algoname \ towards reducing hallucinations in KG-grounded dialogue systems on the OpenDialKG dataset \cite{moon2019opendialkg}.  At present, OpenDialKG is the only publicly available dataset that provides open-ended dialogue responses grounded on paths from a given KG, this is why we limit our experiments on this dataset.
As there are no established metrics for this task, we consider a suite of task-specific and automated metrics to assess the different components of $\algonameshort$ and the degree of hallucination present. We use standard classification metrics such as F1-score, precision and recall to evaluate $C$ and PPL to measure the quality of the LM. Similarly, we use retrieval metrics like Hits@$k$, Mean Rank (MR), and Mean Reciprocal Rank (MRR) to evaluate the Entity Mention Retriever. Precise implementation details can be found in~\S\ref{app:implementation_details}.

\begin{table}[t]
\centering
\begin{small}
\begin{tabular}{ l | c | c | c }
 \hline
  {\small \textbf{Model}} & {{\small \textbf{FeQA}}} & {{\small \textbf{Critic}}} & {{\small \textbf{BLEU}}}  \\
 \hline
 {\small GPT2-KG} & {\small 26.54} & {\small 19.04} &  \small \textbf{11.79*}   \\
 {\small \ \ + \textsc{\algonameshort}} & {\small \textbf{28.98*}} & {\small \textbf{11.72*}} &{\small 11.29}  \\
 {\small \ \ + \textsc{\algonameshort-w/o nce}} & {\small 26.02} & {\small 17.91}  & \small 10.98 \\
  {\small \ \ + \textsc{\algonameshort-w. CompGCN}} & {\small 26.89} & {\small 15.41} & \small 11.10 \\
 {\small \ \ + \textsc{\algonameshort-w/o MLM}} & {\small 27.01} & {\small 15.02}  & {\small 10.88}  \\
  {\small \ \ + \textsc{\algonameshort-w/o critic}} & {\small 18.23} & {\small 19.65} & \small 6.49  \\
 
 \hline
 {\small AdapterBot} & {\small 23.11} & {\small 26.68} &  \small{10.56}  \\
 {\small \ \ + \textsc{\algonameshort}} & {\small \textbf{27.21*}} & {\small \textbf{18.51*}} &  \small \textbf{10.74}* \\
  {\small \ \ + \textsc{\algonameshort-w/o nce}} & {\small 24.02} & {\small 25.02} & \small  9.98  \\
   {\small \ \ + \textsc{\algonameshort-w. CompGCN}} & {\small 25.83} & {\small 20.23}  & \small 10.11 \\
  {\small \ \ + \textsc{\algonameshort-w/o MLM}} & {\small 26.02} & {\small 21.04} & \small 10.06 \\
  {\small \ \ + \textsc{\algonameshort-w/o critic}} & {\small 16.21} & {\small 27.22} &  \small 5.64  \\
 
 \hline
  {\small GPT2-KE} & {\small 19.54} & {\small 28.87 } &  {\small \textbf{6.24}*} \\
  {\small \ \ + \textsc{\algonameshort}} & {\small \textbf{26.21*}} & {\small \textbf{20.34*}} & {\small 6.06} \\
  {\small \ \ + \textsc{\algonameshort-w/o nce}} & {\small 20.34} & {\small 24.32} & \small 5.89 \\
   {\small \ \ + \textsc{\algonameshort-w. CompGCN}} & {\small 23.23} & {\small 21.21} &  \small 6.01   \\
  {\small \ \ + \textsc{\algonameshort-w/o MLM}} & {\small 24.01} & {\small 22.40} &  \small  5.99 \\
  {\small \ \ + \textsc{\algonameshort-w/o critic}} & {\small 15.89} & {\small 30.71} & \small  3.49 \\
  
  \hline
  {\small Gold response} & {\small 33.34} & \small 5.2 &  - \\
\hline
\end{tabular}
\end{small}
\caption{{Measuring the degree of hallucination of different models pre and post-refinement on generated samples based on the OpenDialkg test data.
\cut{FeQA measures the faithfulness score as the average F1
between the answers produced from the generated responses and the answers produced from the dialog history.} A higher FeQA score indicates an increase in faithfulness. The hallucination Critic (Critic) measures the percentage of hallucinated responses in the dataset.}  (\textbf{*} $p$-value $< 0.001$). $ \algonameshort$ uses GPT2 emb. for the KG-Entity Memory.  
}
\label{tab:fix_resp_ok}
\vspace{-5mm}
\end{table}

\xhdr{Hallucination Metrics} We consider $3$ different hallucination metrics \textbf{M1-M3} that provide a multi-faceted measure of performance. Appendix~\S\ref{app:hallu_met} outlines these metrics in detail. Succinctly, \textbf{M1.} BLEU \cite{papineni2002bleu} \cut{which we use as a weak proxy for the faithfulness of a response.} \textbf{M2.} Hallucination Critic which we reapply to the refined response. For \textbf{M3.} we repurpose the FeQA measure \cite{durmus2020feqa}---known to be an effective faithfulness measure in text summarization---by considering our document as the concatenation of $\mathcal{D}$ and all $\mathcal{G}^1_c$ triples while the summary is the response $\bar{x}_{n+1}$. 

\xhdr{Negative Candidates}
We consider two different negative sampling strategies in order to compute $\mathcal{L}_{\text{NCE}}$: SANS \cite{ahrabian2020structure} and In-batch-negatives. SANS selects hard negatives by leveraging the graph structure and selecting negative samples from a context entity’s $k$-hop subgraph (e.g. $\mathcal{G}^1_c$).
Meanwhile, In-batch-negatives considers the ground truth triple of each sample within a batch as a negative candidate for the other
samples in the same batch. Using this approach, the number of candidates is equal to the batch size.

\subsection{Main Experimental Questions}
\label{exp:questions}
Our experiments answer the following questions: 
\begin{itemize}[noitemsep]
  \item[\textbf{Q1)}] \textbf{Identifying Hallucinations.} Can $C$ identify both extrinsic and intrinsic hallucinations?
 \item[\textbf{Q2)}] \textbf{Reducing Hallucinations.} Is $\algonameshort$ effective in reducing hallucinations?
 \item[\textbf{Q3)}] \textbf{Query Generation.} Can \algonameshort \ retrieve the correct entities and is $\mathcal{L}_{\text{NCE}}$ important to learn query representations $q_i$?
 \item[\textbf{Q4)}] \textbf{Impact of MLM and Critic.} Is MLM essential to our training strategy or can we only use  an autoregressive LM? Analagously, can we simply bypass the critic during refinement?
 \item[\textbf{Q5)}] \textbf{Impact of global graph structure.} Is the global graph structure important for learning KG-Entity memory representations? 
 \cut{
  \item[\textbf{Q6)}] \textbf{Impact of the hallucination critic}. Does the critic has an essential role in NPH in improving the performance? What will happen if we bypass the hallucination critic when attempting to fix the hallucinated sentences? 
}
\end{itemize}

\subsection{Results}
\label{baselines}

Throughout our experiments, we rely on three representative baselines for response generation: GPT2-KG, AdapterBot \cite{madotto2020adapter}, and GPT2-KE \cite{madotto2020learning}. GPT2-KG is a small pre-trained GPT2 model \cite{radford2019language} fine-tuned on the dialogue corpus. AdapterBot uses a fixed backbone conversational model such as DialGPT~\cite{zhang2020dialogpt} and encodes multiple dialogue skills via different adapters~\cite{houlsby2019parameter}. Both GPT2-KG and AdapterBot process inputs by concatenating $\mathcal{D}$, $\mathcal{K}_n$ and the generated response. GPT2-KE on the other hand uses a GPT2 model trained on a knowledge-augmented training set. 

\begin{table*}[t]
\begin{small}
\centering
\resizebox{\textwidth}{!}{
\begin{tabular}{ c c | c  c  c  c  c  c  c}
 \hline
   & {\small \textbf{Model}} & {{\small \textbf{Neg. candidates}}} &  {{\small \textbf{PPL}}} &  {{\small \textbf{Hits@1}}} & {{\small \textbf{Hits@3}}} & {{\small \textbf{Hits@10}}} & {{\small \textbf{MR}}} & {{\small \textbf{MRR}}} \\
  \hline
 \multirow{5}{*}{\rotatebox[origin=c]{90}{\scriptsize \texttt{\textbf{GPT2-Emb}}}} & \cellcolor{Gray} & \cellcolor{Gray} {\small SANS } & \cellcolor{Gray}\textbf{8.56} & \cellcolor{Gray}{\small \textbf{0.73}} & \cellcolor{Gray}{\small \textbf{0.92}} & \cellcolor{Gray}{\small \textbf{0.99}} &\cellcolor{Gray} {\small \textbf{1.76}} \cellcolor{Gray}& {\small \cellcolor{Gray}\textbf{0.83}} \\ 
  & \cellcolor{Gray} \multirow{-2}{*}{{\small \algonameshort}} & \cellcolor{Gray}{\small{In-Batch Negatives }} &\cellcolor{Gray} \small \small 8.67  & \cellcolor{Gray}\small 0.42 & \cellcolor{Gray}\small 0.75 & \cellcolor{Gray}\small 0.94 &\cellcolor{Gray} \small 3.08 &\cellcolor{Gray} \small 0.68  \\
  & \multirow{1}{*}{\small \small \textsc{\algonameshort-w/o nce}} & - & \small 9.64 & \small 0.02  & \small 0.05   & \small 0.1  & \small 35.49 &  \small 0.07
  \\
  & \cellcolor{Gray} & \cellcolor{Gray} {\small SANS } & \cellcolor{Gray} \small{9.73}  & \cellcolor{Gray} {\small 0.47} & \cellcolor{Gray} {\small 0.76}  & \cellcolor{Gray} {\small 0.96} & \cellcolor{Gray} {\small 2.83} & \cellcolor{Gray} {\small 0.64}
  \\ 
  & \cellcolor{Gray} \multirow{-2}{*}{\small \cellcolor{Gray} \textsc{\algonameshort-w/o mlm} } & \cellcolor{Gray}  \cellcolor{Gray} {\small{In-Batch Negatives }} & \cellcolor{Gray} \small  9.70 & \cellcolor{Gray} \small 0.20  &\cellcolor{Gray} \small  0.43& \cellcolor{Gray} \small 0.75 & \cellcolor{Gray} \small 9.22 & \cellcolor{Gray} \small 0.36 \\
 \multirow{5}{*}{\rotatebox[origin=c]{90}{\tiny \texttt{\textbf{CompGCN-Emb}}}} & \multirow{2}{*}{\small \small \algonameshort} &  {\small SANS } & \small \textbf{8.99} & \small \textbf{0.13} & \small \textbf{0.26} & \small \textbf{0.52} & \small \textbf{14.27} & \small \textbf{0.25}
  \\ & &  {\small{In-Batch Negatives }}& \small 10.04 & \small 0.08 & \small 0.17 & \small 0.43 & \small 15.75 & \small 0.16  \\
 & \cellcolor{Gray}\multirow{1}{*}{\small \textsc{\algonameshort-w/o nce}} & \cellcolor{Gray} - & \cellcolor{Gray} \small 10.61 &\cellcolor{Gray} {\small 0.04}  &\cellcolor{Gray} {\small 0.12}   &\cellcolor{Gray} \small 0.27  & \cellcolor{Gray}\small 26.50 & \cellcolor{Gray} \small 0.12
  \\
  & \multirow{2}{*}{\small \small \textsc{\algonameshort-w/o mlm} } & {\small SANS } & \small 9.63  & \small 0.08 & \small 0.21  & \small 0.47  & \small  15.52 & \small 0.20
  \\ 
  & & {\small{In-Batch Negatives }} & \small 9.64  & \small 0.02 & \small 0.05 & \small 0.16  & \small 80.52 & \small  0.07 \\
  \hline

\end{tabular}
}
\caption{Ablation studies on $\algoname$ on the gold responses from the OpenDialKG test data.}
\label{tab:editor_training}
\vspace{-5mm}
\end{small}
\end{table*}

\subsubsection*{Q1: Identifying Hallucinations}
Analogous to the study conducted in ~\S\ref{sec:hallu}, we ask humans to identify the span of text that is hallucinated w.r.t. to the given triples in $500$ responses generated greedily from GPT2-KG. We report the average Krippendorf's alpha coefficient to be $0.73$ on the annotator responses. Table \ref{tab:critic} outlines our results. To explore the robustness of our corruption strategies as discussed in ~\S\ref{critic}, we fine-tune a large RoBERTa model \cite{liu2019roberta} 
on three different synthetic datasets: (i) \texttt{RoBERTa-Extrin} corresponds to the negative examples crafted using an extrinsic hallucinations, where entity mentions are first extracted using the SpaCy NER tagger \cite{honnibal2017spacy}.  (ii) \texttt{RoBERTa-Intrin} consists of negative examples that contain intrinsic hallucinations. 
(iii) Finally, \texttt{RoBERTa-Intrin-Extrin} corresponds to examples that were either corrupted using an extrinsic or intrinsic strategy but not both simultaneously. For (i) and (ii), the examples are obtained by corrupting the full train OpenDialKG data.
We observe that \texttt{RoBERTa-Intrin-Extrin} achieves the highest F1 ($70.35\%$), compared to the classifiers trained on the first two synthetic datasets. Such a result highlights that our \texttt{RoBERTa-Intrin-Extrin} classifier can indeed detect both kinds of hallucinations and also that our corruption strategies are effective.  
In the rest of the experiments, we take \texttt{RoBERTa-Intrin-Extrin} as the hallucination classifier $C$.

\subsubsection*{Q2: Reducing Hallucinations}
We evaluate the ability of \algonameshort \ in fixing hallucination in generated responses in the three response generation baselines. We also perform ablation for each model using the different components of \algonameshort. We present the results in Table~\ref{tab:fix_resp_ok} which show the degree of hallucination prior to and after applying \algonameshort \ on each response generation method. We find that \algonameshort \ consistently performs favourably in reducing hallucination across FeQA and the hallucination Critic. In particular, we observe that the strongest iteration of each baseline model is the original model paired with the full $\algonameshort$ module. For example, 
in AdapterBot, \algonameshort \  decreases the Critic score by $8.17$ points and increases faithfulness by $6.67$ points on FeQA. With respect to BLEU scores, we observe inconsistent performance across the different baselines with AdapterBot+\algonameshort\ incurring a marginally higher score. While we use BLEU as a proxy for faithfulness, it is still an imperfect measure as it is computed solely between the n-gram overlap between a reference and generated text which neglects the important fact that there is a multitude of different ways to generate a faithful response w.r.t. a KG.
\cut{
We believe that BLEU is not adequate for measuring faithfulness in dialogue systems as it measures solely the n-gram overlap between a reference and the generated text.  However, in most cases, there is more than one correct way to generate a faithful response w.r.t to KG. This means that a correct response may score low on BLEU.}


\subsubsection*{Q3: Query Generation}
We now investigate \algonameshort's ability to retrieve the correct entity using the crafted query. We present the results in Table \ref{tab:editor_training} along with different ablation studies. We find that key metrics such as Hits@3 and Hits@10 are nearly saturated when using the complete $\algonameshort$ module with GPT2 embeddings for the KG-Entity memory. Furthermore, we notice that all retrieval metrics drop dramatically (e.g.$\downarrow70$ Hits@1 ) when $\mathcal{L}_{\text{NCE}}$ is omitted. Finally, we observe that SANS negatives lead to lower perplexity and better retrieval performance across the board. This is unsurprising since local negative samples are known to be harder and thus provides a richer learning signal  \cite{ahrabian2020structure}. 

\begin{table}[t]
\centering
\begin{tabular}{ l | c | c | c}
 \hline
  {\small \textbf{Model}} & {{\small \textbf{Precision}}} & {{\small \textbf{Recall}}}   & {{\small \textbf{F1}}}  \\
 \hline
 {\small RoBERTa-Intrin} & {\small 44.9} & {\small 32.54} & {\small 37.73}  \\
 {\small  RoBERTa-Extrin} & {\small 68.65} & {\small 46.94} & {\small 55.76}   \\
 {\small RoBERTa-Intrin-Extrin} & {\small \textbf{83.05*}} & {\small \textbf{61.02*}} & {\small \textbf{70.35*}}   \\
 \hline
\end{tabular}
\caption{{Performance of the hallucination critic on the $500$ human-annotated data (\textbf{*} $p$-value $< 0.001$)
}}
\label{tab:critic}
\vspace{-5mm}
\end{table}

\subsubsection*{Q4: Impact of MLM and Critic}
We now gauge the importance of using MLM and Critic within \algonameshort. To assess the MLM component, we replace each contextual representation $m_i \in M_c$ with randomly initialized values. We highlight our findings in Table \ref{tab:fix_resp_ok} where \textsc{\algonameshort-w/o MLM} performs worse than \algonameshort \ across all models. Investigating further in Table \ref{tab:editor_training}, we observe that performance without MLM degrades substantially (e.g. $\downarrow26$ Hits@1) when using pre-trained GPT2 embeddings as entity memory and similarly for CompGCN embeddings. These findings suggest that MLM facilitates the learning of rich masked representations that are useful in downstream applications, a fact which is in line with other works that leverage MLM \cite{roberts2020much, devlin2019bert,joshi2020spanbert}. 
To judge the impact of the critic, we mask out all entity mentions as opposed to only masking out potential hallucinated ones during refinement. In Table \ref{tab:fix_resp_ok}, we find that \textsc{\algonameshort-w/o critic} performs the worst in every metric compared to all baselines which underlines that simply masking all entities---hallucinated or otherwise---in a response is not a productive strategy for effective refinement. 

\subsubsection*{Q5: Impact of global graph structure}
We now investigate the representation of entities in our KG-Entity Memory. We explore two variants: 1) Initializing embeddings as the output of a pre-trained GPT-2 model. 2) Utilizing node embeddings learned by a CompGCN network trained on a standard relation prediction task over the entire graph $\mathcal{G}$. In both these approaches, the embeddings are updated throughout training using \eqref{eq:loss}. As per Table~\ref{tab:editor_training}, we notice a dramatic difference in both perplexity and retrieval performance in favour of using simply the output of a pre-trained GPT-2 model. Such a result may be reconciled by noticing that any specific turn in dialogue local information (e.g. previous turn)---as conversation topics may drift---is significantly more important to generate a faithful response. Thus, enriching entity embeddings with global structure in $\mathcal{G}$ is less beneficial than aligning $\mathcal{G}^k_c$ with the representation space of the autoregressive LM, which for us is also GPT2.

\begin{table}[t]
\begin{small}
\centering
\begin{tabular}{ l | c | c }
 \hline
  {\small \textbf{Model}} & {{\small \textbf{Hallucination}}} & {{\small \textbf{Fluency}}}     \\
 \hline
 {\small GPT2-KG} & {\small 97.5 $\pm$ 0.6} & {\small 92.5 $\pm$ 1.6}   \\
 {\small  GPT2-KG (+ \textsc{\algonameshort})} & {\small \textbf{56.5} $\pm$ 1.2} & {\small 88.5 $\pm$ 0.7}   \\
  \hline
 {\small AdapterBot} & {\small 95.5 $\pm$ 0.8} & {\small 90.5  $\pm$ 0.4}   \\
 {\small AdapterBot (+ \textsc{\algonameshort})} & {\small \textbf{59.0}  $\pm$ 0.5} & {\small 87.5  $\pm$ 1.2}   \\
 \hline
 {\small GPT2+KE} & {\small 97.0  $\pm$ 0.2} & {\small 91.5  $\pm$ 0.7}   \\
 {\small GPT2+KE (+ \textsc{\algonameshort})} & {\small \textbf{58.5}  $\pm$ 0.6} & {\small 86.0  $\pm$ 0.9}   \\
 \hline
\end{tabular}
\caption{{Human Evaluation on $1200$ responses ($200 \times 6$) from different response generation baselines.}}
\label{tab:human_eval_hall}
\vspace{-5mm}
\end{small}
\end{table}

\subsection{Human Evaluation}

In addition to the automated hallucination metrics, we conduct human evaluation to assess \algonameshort's ability to reduce hallucination. We provide human annotators with $200$ hallucinated responses per baseline(~\S\ref{baselines}) as identified by our hallucination critic~\S\ref{critic}. The faithfulness of each response is evaluated by $3$ humans who are provided $\mathcal{D}$, $\mathcal{K}_n$, and the retrieved path from $\mathcal{G}^k_c$. 
We further request annotators to evaluate the fluency of the responses before and after refinement. Results are depicted in Table \ref{tab:human_eval_hall}. 
We see that the hallucination critic achieves a precision of $97.5\%$ for GPT2-KB responses, $95.5\%$ for AdapterBot and $97.0\%$ for GPT2-KE. In contrast, generation methods when paired with \algonameshort \ reduce hallucinations by a large margin $42.05\%$ for GPT2-KB responses with a marginal drop in fluency ($4.32\%$). We also observe similar performance gains for responses generated from AdapterBot and GPT2-KE. \cut{Further details on the evaluation process can be found in~\S\ref{human_an}}

\cut{
Since hallucinated responses and faithful responses are not evenly distributed in the generated responses as indicated in Table \ref{tab:human_assess_gpt2} (hallucinated responses tend to be fewer than faithful ($21.32\%$ v.s $73\%$), we provide humans with examples that are only hallucinated to evaluate the performance of $\algonameshort$ in fixing factual errors. To identify the hallucinated examples, we use the hallucination critic, described in Sec~\ref{critic}, at a sentence-level.  Examples are generated by our response generation baselines (Sec~\ref{exp:questions}) on the test OpenDialKg data. From each model, we pick 200 hallucinated responses identified by the critic. We crowed-sourced human evaluation using the Appen platform where each example was evaluated by 3 annotators. To judge each response, humans were given a dialogue history, the gold triples and the retrieved path from the graph. They were asked to judge whether the responses are hallucinated w.r.t to the triples. They were further asked to evaluate the fluency of the responses before and after getting fixed. Results are depicted in Table \ref{tab:human_eval_hall}. The sentence-level hallucination critic achieved a precision of $97.5\%$ for GPT2-KB responses, $95.5\%$ for AdapterBot and $97.0\%$ for GPT2-KE, showcasing the efficiency of the critic to identify hallucination. 
Results show that NPH reduces hallucination by a large margin $42.05\%$ for GPT2-KB responses without sacrificing too much the fluency of the sentence which decreases by $4.32\%$. The same observation is made for responses generated from AdapterBot and GPT2-KE.}

\section{Related Work}
\xhdr{Knowledge Graphs} 
 Building large-scale repositories of knowledge has been one of the principle
directions of research in artificial intelligence since
the inception of the field \cite{newell1956logic, newell1959report}. Often represented as large
scale multi-relational graphs, KGs
have seen wide application in a variety of domains,
such as question answering \cite{yao2014information, hao2017end}, and natural language processing \cite{berant2013semantic, yu2014improving} to name a few. Beyond academic research, public
KG’s like FreeBase \cite{bollacker2008freebase} have
been invaluable in industrial applications forming
symbolic backbones of most important products \cite{singhal2012introducing}. KG’s have also risen in prominence in the context of dialogue models that propose to explicitly embed symbolic knowledge representations into a neural embedding space \cite{liu2019roberta, zhu2017flexible, moon2019opendialkg, zhou2018commonsense, xu2020user}. \citet{niu2019knowledge} use a knowledge retriever component that
conditions the response by retrieving relevant facts
from the KG based on the current utterance. Similarly, \citet{young2018augmenting} and \citet{zhou2018commonsense} use a
commonsense KG to inject commonsense knowledge into the response of the conversational model.
\citet{tuan2019dykgchat} explore the effects of using a
dynamic KG in the dialogue model. On the other hand, \citet{moon2019opendialkg}  propose a conversational
reasoning model that traverses a large scale KG to
retrieve a relevant path given a starting node and a classifier to predict the next node a response show
follow. Unlike the KG path traversal problem, 
 this work focuses on removing hallucinations
in generated responses using a KG.

\xhdr{Hallucination} The injection of false information is a well-known phenomena in data-to-text generation~\cite{tian2019sticking,dhingra2019handling,parikh2020totto}, machine translation~\cite{koehn2017six,lee2018hallucinations}, image captioning~\cite{rohrbach2018object}, machine summarization \cite{maynez2020faithfulness, durmus2020feqa} \cut{exposure bias ~\cite{wang2020exposure},} and question answering~\cite{feng2018pathologies}. In the context of dialogue systems, \citet{duvsek2018findings,duvsek2020evaluating} demonstrate that state-of-the-art natural language generation (NLG) models can hallucinate by missing important entities. Few NLG models have been proposed to cope with the issue, but are often custom-made for task-oriented dialogue~\cite{balakrishnan2019constrained}. 
Recently, little progress has been made for studying hallucination in open-domain dialog systems. \citet{dziri2021evaluating} study hallucination in knowledge-grounded dialogue systems and introduce a the BEGIN benchmark for measuring groundedness in dialogue systems.\cut{which consists of $8$K generated dialogue turns with accompanying human annotations. This benchmark serves as a testbed for evaluation metrics built for measuring groundedness in dialogue systems.} Finally, \citet{rashkinincreasing} propose a dialogue system that is more faithful to the source knowledge by adding control tokens at training time that guide the model towards generating more objective sentences which have higher overlap with the source.

\cut{Beyond academic research, public KG's like FreeBase \cite{bollacker2008freebase} have been invaluable in industrial applications forming symbolic backbones of most important products \cite{singhal2012introducing}. }

\cut{\citet{liu2019knowledge} use a knowledge retriever component that conditions the response by retrieving relevant facts from the KG based on the current utterance. Similarly, 
\citet{young2018augmenting,zhou2018commonsense} use a commonsense KG to inject commonsense knowledge into the response of the conversational model. \citet{tuan2019dykgchat} explore the effects of using a dynamic KG in the dialogue model.}
\section{Conclusion}

In this work, we investigate the open problem of
hallucination in KG-grounded dialogue systems
and demonstrate that these models are more susceptible to extrinsic hallucinations which predominantly manifest as the injection of erroneous entities. To tackle this challenging problem, we propose a new module $\algoname$ that
aims to enforce faithfulness in KG-grounded dialogue systems by identifying and refining hallucinations via queries over a k-hop subgraph.  We empirically observe that NPH
is capable of reducing hallucination when paired
with a number of base dialogue models with relative improvements of 20.35\% over vanilla GPT2 on FeQA. Our
findings also reveal the crucial role the representation of the local subgraph plays as external memory compared to the full global graph. In this work, we considered a paired KG aligned with dialogue but in many other applications, such dialogue
to KG alignment may be difficult to easily obtain
necessitating the usage of the full graph which is interesting direction
for future work.


\section*{Acknowledgements}
We are grateful to the anonymous
reviewers for helpful comments. This research is
supported by Natural Sciences and Engineering research Council of Canada, the Alberta Machine Intelligence Institute Fellow Program and the Canadian Institute for Advanced Research AI Chair Program. This research is also supported in part by Compute Canada.

\clearpage
\bibliography{anthology,custom-v1}
\bibliographystyle{acl_natbib}

\clearpage
\appendix 
\section{OpenDialKG}
\label{appendix:opendial_kg}

We use OpenDialKG \cite{moon2019opendialkg}, a crowded-sourced English dialogue dataset where two workers are paired together to chat about a certain topic. The first speaker is asked to initiate the conversation about a given entity and the second speaker is tasked to form a factual response based a set of facts extracted from an existing KG, Freebase \cite{bast2014easy}. Those facts represent paths in the KG that are either 1-hop or 2-hop from the initial entity. Once the second speaker sends a response, the first speaker continues discussing the topic engagingly and new multi-hop facts from the KG are presented to the second speaker.
The conversation can be regarded as traversing multiple paths in the KG. However, not all utterances within the same dialogue are grounded on facts from the KG. The second speaker can choose not to select a path from the KG to form an answer and instead forms a ``chit-chat" response.
Overall, the dataset consists of four domains: movie, music, sport and book where each second speaker's utterance is annotated with paths from the KG. The KG corresponds to a large subgraph extracted from Freebase with $\sim1.2$M triples (subject, predicate, object), $\sim101$k distinct entities and 1357 distinct relations. No official
split is provided in the original dataset, and thus we randomly split the dataset in 80/10/10 for the train/valid/test, respectively.
The data consists of 61778 train, 7933 valid and 7719 test.   Some utterances in the dataset are chit-chat and thus are not annotated with a path from the KG. Thus, we filter the dataset by keeping only the dialogue examples that are annotated with a path from the KG. We ended up with 23314 training examples, 2954 valid examples and 2954  test examples.

\section{Human Evaluation for Modes of Hallucination}
\label{detailed_mode_hall}
 Workers, fluent in English, were trained for the task before starting the evaluation process. If the workers fail to achieve at least 80\% accuracy in answering the different test questions, they would not be allowed to start the evaluation process. These workers are hired from Appen \footnote{\url{https://appen.com/}}.
 Each worker was presented with a dialogue history, knowledge triples including the gold triples and the 1-hop paths from the centre node in $\mathcal{G}^k_c$. Each example was evaluated by 3 workers and majority vote was considered. 
 
 Workers were asked the following questions:
 \begin{enumerate}
     \item Is this response hallucinated with respect to the gold knowledge triples? (Most definitely, Not at all)
     \begin{enumerate}
         \item If the response is hallucinated, does it represent extrinsic hallucination, intrinsic hallucination or both? (Extrinsic, Intrinsic, Both)
        
     \end{enumerate}

    \item  If the response is not hallucinated, is it faithful to the source or generic? (Faithful, Generic)
    
       \item Is this a coherent response with respect to the dialogue history even if it was identified as hallucinated? (Most definitely, Not at all)
    
    \item Is this response grammatically correct? (Most definitely, Not at all)

 \end{enumerate}

 \section{KG-Entity Memory}
\paragraph{GPT2 embeddings} OpenDialKG contains a textual description, called ``render'', for triples extracted from the KG. Note that not all triples in the dataset are associated with ``render''. To get a contextual representation for each entity mention, we feed ``render'' to GPT2 and then  extract hidden states representations for each entity's  word piece and finally obtain a final representation by applying a MaxPool over the hidden representations. For entity mentions that are not described in ``render'', we get their representations directly from the last hidden states in GPT2.

  \begin{table*}[t]
\centering
\begin{tabular}{ l | c c c | c | c | c | c}
 \hline
  \multirow{2}{*}{\small \textbf{GPT2}} & \multicolumn{3}{c|}{{\small \textbf{Hallucination}}} & {{\small \textbf{Faithfulness}}}   & {{\small \textbf{Generic}}}  & {{\small \textbf{Coherence}}} & {{\small \textbf{Fluency}}} \\
  & \small \textbf{Ex} & \small \textbf{In} & \small \textbf{B} & & & \\
 \hline
 {\small Greedy} & {\small 17.66 $\pm$ 2.6} &  {\small 2.00  $\pm$ 3.5}&{\small 1.66  $\pm$ 0.5} & {\small 69.00 $\pm$ 3.2} & {\small 9.66  $\pm$ 2.7 } & {\small 81.66  $\pm$ 3.2 } & {\small  95.67  $\pm$ 1.6}\\
 {\small Beam Search} & {\small 18.33  $\pm$ 2.8} & {\small 3.33  $\pm$ 3.8} & {\small 4.00  $\pm$ 1.8} & {\small 68.00  $\pm$ 3.9} & {\small 6.33  $\pm$ 2.7 } &  {\small 83.33  $\pm$ 1.6} & {\small 97.00  $\pm$ 1.9}\\
 {\small Nucleus 0.9} & {\small 25.33  $\pm$ 2.1} & {\small 4.00  $\pm$ 3.6} &{\small 2.33  $\pm$ 3.6} & {\small 64.66  $\pm$ 2.3}& {\small 3.66 $\pm$ 3.2} & {\small 83.66  $\pm$ 2.4}  & {\small 99.10  $\pm$ 0.6} \\
  {\small Nucleus 0.5} & {\small 23.33  $\pm$ 2.2} & {\small 5.33  $\pm$ 3.1} & {\small 4.33  $\pm$ 0.8} & {\small 59.90 $\pm$ 2.5} & {\small 7.00 $\pm$ 2.6} & {\small 87.66  $\pm$ 2.1}  & {\small 98.34  $\pm$ 0.4} \\
  {\small Top20} & {\small 28.33  $\pm$ 1.5} & {\small 7.00  $\pm$ 2.6}& {\small 5.00  $\pm$ 1.5} & {\small 55.00  $\pm$ 0.6} & {\small 4.66  $\pm$ 1.8} &  {\small 80.33  $\pm$ 1.6} &{\small 97.34  $\pm$ 0.5} \\ \hline
\end{tabular}
\caption{{\small Human assessment of random 1500 GPT2 dialogue responses (300 $\times$ 5) generated based on the test OpenDialkg data \cite{moon2019opendialkg}(mean preferences $\pm$90\% confidence intervals). 
}}
\label{tab:human_assess_gpt2_appe}
\end{table*}

 

\section{Implementation Details}
\label{app:implementation_details}
\paragraph{NPH:}
\algonameshort \ is implemented using the Pytorch Huggingface Transformers library \cite{wolf-etal-2020-transformers} and  the Pytorch-lightning library \cite{falcon2019pytorch}. Concretely, we use a small RoBERTa model \cite{liu2019roberta} as the MLM and the base GPT2 model \cite{radford2019language} as our autoregressive LM. During training, we use the Adam optimizer \cite{kingma2015adam} with Dropout \cite{srivastava2014dropout} on a batch size of $16$ with a learning rate of $6.25 \times 10^{-5}$ that is linearly decayed. The maximum dialogue history length is set to $3$ utterances. The coefficient $\lambda$ in \eqref{eq:loss} is set to $0.5$.  We varied the factor from 0.1 to 1 and 0.5 was chosen based on the best results on the validation set. The number of negative examples is set to $50$ for SANS. The model early-stops at epoch 10 and we save the best model based on the validation set. Our hyperparameters search is done via greed search. The average runtime of this model is 4 hours.

\paragraph{GPT2-KG:}Similarly, we implement this baseline using the  Pytorch Huggingface Transformers library \cite{wolf-etal-2020-transformers}  and  the Pytorch-lightning library \cite{falcon2019pytorch}. During training, we use the Adam optimizer \cite{kingma2015adam} with Dropout \cite{srivastava2014dropout} on a batch size of $32$ with a learning rate of $6.25 \times 10^{-5}$ that is linearly decayed. The maximum dialogue history length is set to $3$ utterances. The model early-stops at epoch 6.  The average runtime of this model is 2 hours.

\paragraph{AdapterBot and GPT2-KE:} We use the code that's  publicly available by the authors at \url{https://github.com/HLTCHKUST/adapterbot} and \url{https://github.com/HLTCHKUST/ke-dialogue} and we follow closely their training procedure described in \citep{madotto2020adapter} and \citep{madotto2020learning}. We use the GPT2-KE with 9K iterations. The average runtime of these models is 3 hours.

Training for all models, including baselines, is done on an Nvidia V100 GPU 32GB and for inference, we use greedy search. 

\paragraph{Hallucination Critic:} We use a pre-trained RoBERTa-large  classifier  \cite{liu2019roberta} provided by the Huggingface Transformers library \cite{wolf-etal-2020-transformers}. The model was trained using the Adam optimizer with a learning rate of $2 \times 10^{-5}$ for $5$ epochs on one Nvidia V100 GPU 32GB.  The average runtime of this model is 2 hours.

\section{Hallucination Metrics}
\label{app:hallu_met}
Although BLEU measures the extent to which the generated response is similar to the reference faithful response, it can be misleading in the case where the generated response is very distant from the ground-truth response but faithful to the knowledge triples.
We consider 2 other metrics that focus on measuring the degree of hallucination in the generated responses:

\paragraph{Hallucination Critic} We use our trained token-level hallucination critic as a sentence-level hallucination detector. We consider the utterance as hallucinated if at least one token was identified as hallucinated. As input, the critic receives the dialogue history, the gold triples and the generated response and the output is a binary label indicating hallucination or not. To use this critic for the output of $\algonameshort$, we augment the gold triples with the path extracted based on the Entity Mention Retriever.

\paragraph{FeQA} \citet{durmus2020feqa} has been shown successful in measuring faithfulness in the text summarization task. It generates questions from the candidate summaries and then answers them against the input documents. It measures the average F1 score against the gold answers from the document. Through asking and answering questions, FeQA measures the semantic correctness of the generated responses. To adapt FeQA to our dialogue task, we flatten each path into a pseudo sentence by joining the $\langle \texttt{[SBJ]}, \texttt{[PRE]}, \texttt{[OBJ]} \rangle$ with a simple space, e.g., [Crescendo, written by, Becca fitzpatrick] $\rightarrow$ ``Crescendo written by Becca Fitzpatrick". We consider our document as the concatenation of $\mathcal{D}$ and all $\mathcal{G}^1_c$ triples and the candidate summary as the generated/refined response. FeQA takes a given generated grounded response as input, and generates
questions. It then employs a QA system to answer the generated questions based on the knowledge the response was grounded in.

We use the code made publicly available by the authors \footnote{\url{https://github.com/esdurmus/feqa}}. A similar work to FeQA is QAGS \cite{wang2020asking} which corresponds to asking and answering questions to evaluate the factual consistency of summaries.
\section{Human Evaluation of NPH responses}
\label{human_an}
Analogous to evaluating modes of hallucination, we solicit human evaluation from Appen \footnote{\url{https://appen.com/}} where we train English-speaking annotators for the task before starting the evaluation process.  To evaluate the responses generated by our response generation baselines, annotators were presented with  $\mathcal{D}$, $\mathcal{K}_n$ and the generated response. And, to evaluate NPH's responses, annotators were presented with  $\mathcal{D}$, $\mathcal{K}_n$, the retrieved path from $\mathcal{G}^k_c$ and the refined response. Humans were asked to answer the following questions:
\begin{enumerate}
     \item Is this response hallucinated with respect to $\mathcal{K}_n$? (Most definitely, Not at all)
    
    \item Is this a fluent response, i.e., a response that's grammatically correct? (Most definitely, Not at all)

 \end{enumerate}
 
 In total, humans evaluated 1200 responses: 600 responses (200 from each response generation baseline before refinement) and 600 responses after refinement. 

\section{Error Analysis}
To gain insight into the potential shortcomings of $\algoname$, we conduct an error analysis on refined responses that still contain undesirable hallucinations. Examples of failed refinements using \algonameshort \ are listed below. Recall that for effective retrieval \algonameshort \ requires oracle access to $\mathcal{G}^k_c$ which pre-supposes  the existence of the correct entity in the subgraph. However, based on the examples below, we observe that many of the failed retrievals correspond to entities that might exist in $\mathcal{G}$ but are critically not supported within $\mathcal{G}^k_c$. To highlight this point, let us consider the following example:

\cut{
Recall, we are using the subjects and objects of the gold triples to obtain the subgraph which we query to retrieve the correct entity mentions. We noticed that the generated response is sometimes hallucinating about an entity mention that is not within the 1-hop radius of the gold triples. Consider the following example:}
\begin{quote}
        \footnotesize{\textsf{\textbf{Previous turn}: Could you recommend a book similar to Thirteen Reasons Why?}}
    \end{quote}
     \begin{quote}
        \footnotesize{\textsf{\textbf{Gold triple}:[['Thirteen Reasons Why', 'has genre', 'Young-adult fiction'] }}
    \end{quote}
      \begin{quote}
        \footnotesize{\textsf{\textbf{GPT2-KB Response}: Sure, there is a book called \textbf{The Sea of Monsters} by \color{red}{John Green}.}}
        \end{quote}
         \begin{quote}
        \footnotesize{\textsf{\textbf{Critic}: John Green.}}
        \end{quote}
          \begin{quote}
        \footnotesize{\textsf{\textbf{Context nodes}: Thirteen Reasons Why, Young-adult fiction}}
         \end{quote}
        \begin{quote}
        \footnotesize{\textsf{\textbf{Top-1 Path}:[[Thirteen Reasons Why, written by, Jay Asher]}}
         \end{quote}
           \begin{quote}
        \footnotesize{\textsf{\textbf{NPH response}: Sure, there is a book called The Sea of Monsters written by \color{red}{Jay Asher}.}}
    \end{quote}

In this example, we notice that the original response contains hallucination as ``John Green" did not write the book titled ``The Sea of Monsters''. Moreover, to fix this response $\algonameshort$ has access to $\mathcal{G}^1_c$ which contain the context nodes ``Thirteen Reasons Why'' and ``Young-adult fiction'' but not the entity of interest ``The Sea of Monster''. As a result the proposed fix using \algonameshort \ is the wrong entity ``Jay Asher''. It is important to note that the perceived mistakes by \algonameshort \ is in fact a limitation of our assumption that the correct entities not only exist in $\mathcal{G}$ but can be extracted by running a query on the extracted $k$-hop subgraph. If this fails to be the case in practice, then we cannot hope for \algonameshort \ to retrieve the correct entity. We hypothesize that extending \algonameshort \ to reason over multiple subgraphs or even $\mathcal{G}$ in an efficient manner will serve to alleviate this failure mode.
\cut{
In future work, we aim to enhance NPH with an additional module that focuses on retrieving the relevant subgraph given the history and the generated response. 
}

In addition to using the incorrect subgraph for retrieval, another source of errors is caused by the failure to identify the hallucinated entities to fix in the first place. This can occur in practice as the Critic $C$ is not perfect and may occasionally miss hallucinated entities and thus propagate the error forward. The example below illustrates this failure mode:

\begin{quote}
        \footnotesize{\textsf{\textbf{Previous turn}: Yes, I watched them. Could you tell me more about The Heart Is Deceitful Above All Things, please.}}
    \end{quote}
     \begin{quote}
        \footnotesize{\textsf{\textbf{Gold triple}:[The Heart Is Deceitful Above All Things, starred actors, Jimmy Bennett]}}
    \end{quote}
      \begin{quote}
        \footnotesize{\textsf{\textbf{GPT2-KB Response}: Sure, it was written by {\color{red}{Tony Kaye}} and stars \color{red}{Scarlett Johansson}.}}
        \end{quote}
          \begin{quote}
        \footnotesize{\textsf{\textbf{Critic}: Scarlett Johansson.}}
        \end{quote}
          \begin{quote}
        \footnotesize{\textsf{\textbf{Context nodes}: The Heart Is Deceitful Above All Things, Jimmy Bennett}}
         \end{quote}
        \begin{quote}
        \footnotesize{\textsf{\textbf{Top-1 Path}:[The Heart Is Deceitful Above All Things, starred actors, Jimmy Bennett]}}
         \end{quote}
           \begin{quote}
        \footnotesize{\textsf{\textbf{NPH response}:  Sure, it was written by {\color{red}{Tony Kaye}} and stars \color{emerald}{Jimmy Bennett}}.}
    \end{quote}
    
We believe a stronger Critic or other mechanisms to identify token-level hallucinations can improve the downstream performance of \algonameshort.
\cut{
We plan to fix this issue by training a better Critic and this require better token-leven training data. 
}

The final source for error---while not frequent---is the fluency of the refined response. We find that in an effort to remove hallucinations, other undesirable properties such as a lack of grammaticality may be introduced. The example below demonstrates such a failure mode. This error can be mitigated if we associate each node in the graph by an entity type. 
\begin{quote}
        \footnotesize{\textsf{\textbf{Previous turn}: Sure. Isn't he married to Coleen Rooney}}
    \end{quote}
     \begin{quote}
        \footnotesize{\textsf{\textbf{Gold triple}:[[United Kingdom, Country of nationality, Wayne Rooney]}}
    \end{quote}
      \begin{quote}
        \footnotesize{\textsf{\textbf{GPT2-KB Response}: No, Yes, he is married to \color{red}{Caroline Wozniacki}.}}
        \end{quote}
          \begin{quote}
        \footnotesize{\textsf{\textbf{Critic}: Scarlett Johansson.}}
        \end{quote}
          \begin{quote}
        \footnotesize{\textsf{\textbf{Context nodes}: United Kingdom, Wayne Rooney}}
         \end{quote}
        \begin{quote}
        \footnotesize{\textsf{\textbf{Top-1 Path}:[United Kingdom, Country of nationality, Wayne Rooney}}
         \end{quote}
           \begin{quote}
        \footnotesize{\textsf{\textbf{NPH response}: Yes, he is married to \color{red}{United Kingdom}.}}
    \end{quote}

\section{Sample Refined Responses}
Some examples of the refined responses are presented in Table \ref{tab:examples_fixed}.
\label{app:samples}
 \enlargethispage{2em}
 \begin{table*}[htbp]
 \centering
    \begin{tabu}to\linewidth{@{}X[1.2,l]X[0.2,l]X[6,l]@{}}
    \toprule
 
      \textbf{\small History} & $A$: & I love Shakespeare.  Romeo and Juliet is my favorite.\\
      & $B$: & I like that one as well. The play was written by Kylie Scott.\\
      & $A$: & She also wrote Deep, did she not?\\
      \midrule
      \textbf{\small GPT2-KG} & $B_{gen}$ & Yes she did. She also wrote \colorbox{lightcoral}{The Tragedy of King Richard the Third}. \\
    \midrule
     \textbf{\small Gold knowledge}
    &  $T_1$ & [Deep, written by, Kylie Scot] \\
        \midrule
     \textbf{\small Top-5 Paths}
    & $T'_1$: & [Play,  written by, Kylie Scott]
\\
    & $T'_2$: & [Dirty, written by, Kylie Scott]
 \\
    & $T'_3$: & [Lick, written by, Kylie Scott] \\
    & $T'_4$: & [Deep, written by, Kylie Scott]
 \\
    & $T'_5$: & [Trust, written by, Kylie Scott]
    \\   \midrule
       \textbf{\small $\algonameshort$ response} & $B_{fix}$ &Yes she did. She also wrote \colorbox{emerald}{Play}. \\
    \bottomrule
      \bottomrule
      \textbf{\small History} & $A$: & Do you know of a movie directed by Jean Pierre Jeunet?\\
      \midrule
      \textbf{\small GPT2-KG} & $B_{gen}$ & Jean-Pierre jeunet directed \colorbox{lightcoral}{The Bourne Ultimatum}.\\
    \midrule
     \textbf{\small Gold knowledge}
    &  $T_1$: & [Micmacs, written by, Jean Pierre Jeunet] \\
    &  $T_2$: & [Micmacs, has genre, French]\\
        \midrule
     \textbf{\small Top-5 Paths}
    & $T'_1$: & [Alien: Resurrection, directed by, Jean Pierre Jeunet]
\\
    & $T'_2$: & [A Very Long Engagement, directed by, Jean Pierre Jeunet]
 \\
    & $T'_3$: & [Amélie, directed by, Jean Pierre Jeunet] \\
    & $T'_4$: & [The Extraordinary Voyage, starred actors, Jean Pierre Jeunet]
 \\
    & $T'_5$: & [Micmacs, written by, Jean Pierre Jeunet]
    \\   \midrule
       \textbf{\small  $\algonameshort$ response} & $B_{fix}$ &Jean-Pierre jeunet directed \colorbox{emerald}{Alien: Resurrection}. \\
    \bottomrule
    \end{tabu}
    \caption{Selected responses based on GPT2-KG test responses before and after applying \algoname. The span of texts highlighted in red indicate the hallucinated entity mentions whereas the ones highlighted in green indicate the retrieved correct entity mentions. }
    \label{tab:examples_fixed}
    \end{table*}

\end{document}